\documentclass[]{style/ceurart}
\usepackage{pgfplots}
\usepackage{pgfplotstable}
\usepackage[table]{xcolor}
\usepackage{listings}
\usepackage{placeins}

\begin{document}

\copyrightyear{2026}
\copyrightclause{Copyright for this paper by its authors.
  Use permitted under Creative Commons License Attribution 4.0
  International (CC BY 4.0).}

\conference{CLEF 2026 Working Notes, 21 -- 24 September 2026, Jena, Germany}

\title{DS@GT ARC: Geometric Filtering for Privacy Evaluation of CT Slice Generation}
\title[mode=sub]{ImageCLEF at CLEF 2026}

\author[1]{Eric Regina}[
    orcid=0009-0003-4991-3601,
    email=eric.regina@gatech.edu,
]
\cormark[1]
\author[1]{Richard Arnaud}[
    orcid=0009-0004-2788-2135,
    email=rarnaud3@gatech.edu,
]
\author[1]{Samir {Hadi Cisneros}}[
    orcid=0009-0006-1280-987X,
    email=hadi@gatech.edu,
]

\address[1]{Georgia Institute of Technology, North Ave NW, Atlanta, GA 30332}
\cortext[1]{Corresponding author.}

\begin{abstract}
We present a privacy-focused framework for synthetic lung CT slice generation developed for the ImageCLEFmed GANs 2026 challenge. The approach combines standard Optimal Transport Conditional Flow Matching with a post-generation Supervisor pipeline that filters generated candidates in learned geometric latent spaces using autoencoder embeddings, Determinantal Point Processes, and Stein Kernel Thinning. In the official evaluation, a 100-epoch flow-matching run achieved the highest Privacy Preservation Score among our submissions, with a score of 0.549 and an FID of 0.3290. A geometrically filtered variant achieved the best visual fidelity, with an FID of 0.2639 and a Privacy Preservation Score of 0.492. Geometric filtering generally reduced nearest-neighbor memorization and membership-inference leakage, although its effects varied across the evaluated attacks. Persistent patient re-identification scores indicate that reducing direct similarity to training images is not sufficient to eliminate patient re-identification risk, highlighting an important direction for privacy-focused medical image generation.
\end{abstract}

\begin{keywords}
  Flow Matching \sep
  Synthetic Data \sep
  Medical Imaging \sep
  Subset Selection \sep
  ImageCLEF
\end{keywords}

\maketitle


\section{Introduction}

Artificial Intelligence has the potential to be transformational for medical applications. AI could be applied in the medical industry to more accurately diagnose patients, predict illness to save lives early, provide meaningful categorization of medical data, and ultimately optimize clinical workflows.

While medical AI can be transformational, it requires a significant amount of data to train the models to be effective. But, medical data is highly sensitive and is protected by the Health Insurance Portability and Accountability Act (HIPAA). The process of obtaining useful medical data for each specific domain is an arduous task that requires vast amounts of patient data, strict regulatory approval, and careful anonymization procedures. These challenges have motivated researchers to explore synthetic medical data generation techniques that can generate realistic and diverse medical images for training AI models while preserving patient privacy. Modern generative models, such as Generative Adversarial Networks (GANs) \cite{goodfellow2014generative} and diffusion models \cite{dhariwal2021diffusion}, are capable of producing highly realistic medical images; however, preventing these models from memorizing and reproducing patient-specific anatomical features from their limited training data remains a major challenge.

To tackle this challenge within the context of the ImageCLEFmed GANs 2026 Subtask 3 competition \cite{ImageCLEF2026, ImageCLEFmedicalGANs2026}, our team leveraged Optimal Transport Conditional Flow Matching (OTCFM) \cite{lipman2023flow}, utilizing a 34.5M parameter UNet model. Flow matching offers a highly stable and computationally efficient alternative to traditional diffusion baselines, providing a robust framework for capturing complex, high-fidelity anatomical distributions. 

The objective of this task is to go beyond simple visual replication to manage the trade-off between image realism (measured via a CT-specific Fr\'{e}chet Inception Distance (FID)) and comprehensive privacy preservation of the generated images. The images generated by these models went through an evaluation protocol consisting of four distinct privacy attacks designed by the CLEF task organizers \cite{ImageCLEFmedicalGANs2026} to expose system vulnerabilities: instance-level nearest-neighbor distance (NND) audits, distributional membership inference attacks (MIA), structural patient re-identification, and white-box generalized inversion attacks. 

To protect patient identity across these diverse threat vectors, our team explored multiple different approaches. We investigated multiple inference-time geometric filtering approaches via a post-generation "Supervisor" pipeline.  This pipeline generates a large candidate pool, encodes the generated images using an autoencoder trained on the training data, ranks candidates by their proximity to the training set, and finally applies advanced subset selection techniques such as Determinantal Point Processes (DPP) \cite{kulesza2012determinantal}  and Stein Kernel Thinning (SKT) \cite{dwivedi2021kernel}. Ablations of encoders, distance metrics, and selection methods were applied.

In these working notes, we present a systematic evaluation of our experimental grid. Our submission with the highest PPS score, our attempt at a Conditional Flow Matching framework utilizing a UNet model (\texttt{fm-unet-100}) trained for 100 epochs, had the healthiest balance on the leaderboard. This architecture achieved our highest overall Privacy Preservation Score (PPS) of 0.549 while maintaining an exceptional visual realism score with a Fr\'{e}chet Inception Distance (FID) of 0.3290. Its more favorable privacy results can be attributed to early stopping, which significantly reduced memorization. While our post-hoc filtering successfully reduced image-copying leaks, our broader findings expose a deeper vulnerability regarding distribution-level patient identity retention. We detail this empirical journey and our structural insights to guide future designs in secure medical image generation. The source code for our training pipeline may be found at \url{https://github.com/dsgt-arc/imageclef-med-gans-2026}.

\section{Related Work}

\subsection{Generative Modeling in Healthcare.}
The application of generative models to medical imaging has historically been dominated by Generative Adversarial Networks (GANs) \cite{goodfellow2014generative} and, more recently, Denoising Diffusion Probabilistic Models (DDPMs). While diffusion models generally achieve superior image fidelity and diversity compared to GANs \cite{dhariwal2021diffusion}, they are often characterized by slow, iterative sampling processes. Recently, Optimal Transport Conditional Flow Matching (OT-CFM) has emerged as a highly efficient alternative, bridging the gap between normalizing flows and diffusion processes \cite{lipman2023flow}. By regressing vector fields that map simple base distributions to complex empirical data distributions, Flow Matching enables fast, high-fidelity generation, which is crucial for generating high-resolution medical structures like CT slices.

\subsection{Subset Selection and Geometric Filtering.}
Because generative models often exhibit unequal density coverage or subtle memorization, post-hoc subset selection is frequently employed to curate synthetic datasets \cite{azadi2019discriminator}. Determinantal Point Processes (DPPs) are widely used in machine learning to enforce diversity by modeling the probability of drawing a subset proportional to the volume spanned by its feature embeddings \cite{kulesza2012determinantal}. More recently, Kernel Thinning (KT) and Stein Kernel Thinning (SKT) have been introduced as advanced coreset selection techniques \cite{dwivedi2021kernel}. These methods provide theoretically grounded approaches for summarizing complex probability distributions, ensuring that a selected subset of synthetic images remains both highly diverse and representative of the true data manifold without replicating isolated training outliers.

\section{Methodology}

\begin{figure}[htbp]
    \includegraphics[width=6.25in]{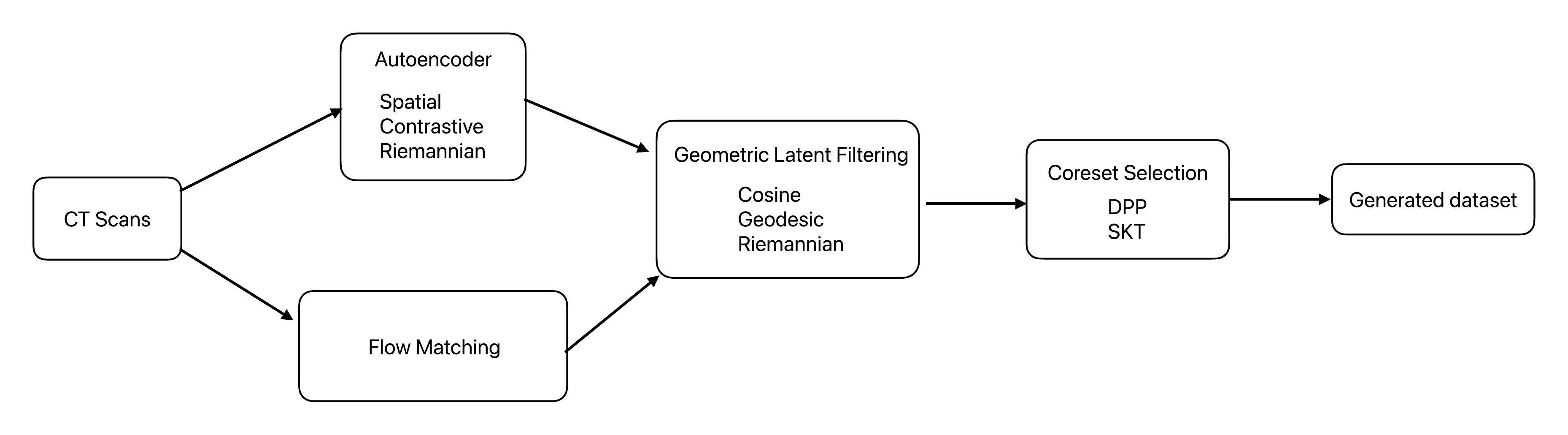}
    \caption{Image generation pipeline overview}
\end{figure}

Our proposed framework addresses the trade-off between clinical realism and patient privacy through a two-stage architecture: the optimization of a continuous normalizing flow generator, followed by a post-generation geometric filtering pipeline, which we refer to as the ``Supervisor''.

\subsection{Dataset and Pre-processing}
The dataset provided for Subtask 3 consists of 10,000 unlabeled 256×256 two-dimensional axial lung CT slices. We identified 69 exact binary duplicates and removed them from the dataset, leaving 9,931 unique slices. We then applied an 80/20 train/holdout split, resulting in 7,945 training images and 1,986 holdout images. 

\begin{figure}[t]
  \centering
  \includegraphics[width=0.9\linewidth]{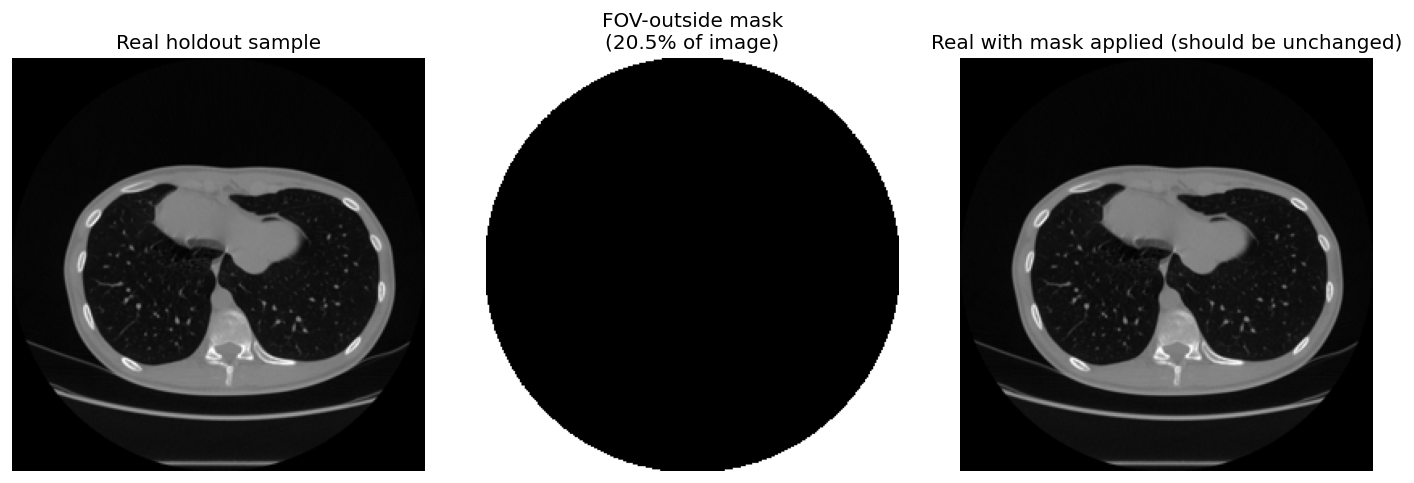}
  \caption{Consensus field-of-view (FOV) mask derived from the real slices.
  A pixel is marked as outside-FOV if it is near-zero in at least 90\% of real
  slices.}
  \label{fig:fov-mask}
\end{figure}

\paragraph{FOV Mask.} Earlier trained versions of our RAM encoder rejected 100\% of generated samples, marking them all as off-manifold regardless of the generator method. The generators ablations used to train RAM were relatively well-trained which led us to believe our encoder had an oversensitivity issue that could potentially be addressed by architecture, objectives, augmentation, and/or pre-processing. For pre-processing, we theorized that each axial slice contained a circular Field of View (FOV) that contained the bulk, if not all of the information we were looking for. To test this theory, we compared spectral and spatial statistics between real and generated slices. Using FFT spectral analysis, we measured the noise floor (pixel variation in the empty corners of each slice) by sampling 32 x 32 pixel patches in the upper left and right corners of each slice. After confirming each patch contained no anatomical data, we found that generated slices contained a noise floor standard deviation approximately twenty times that of real slices ($7.6\times10^{-4}$ vs.\ $3.7\times10^{-5}$). These small perturbations in the background entirely determined the encoder’s ability to distinguish realistic samples from off-manifold samples. Each generated sample carried unnecessary noise that acted as a giveaway, making real and generated samples trivially separable.

To resolve this, when training the Riemannian encoder, we built a circular field-of-view mask from outside-FOV region of real slices. Rather than naively zeroing out all near-zero pixels in our generated slices, we derived a deterministic mask with a soft threshold for natural variation slightly outside of the axial slice FOV. Each pixel was marked as outside-FOV only if it was near zero in 90\% of real slices. We found that applying this consensus FOV mask to generated slices enabled the encoder to evaluate samples based on anatomical content rather than outside-FOV noise. This led to faster and more stable convergence during training.

\subsection{Data Configuration and Generating Model}

To generate high-fidelity synthetic medical images, we utilized Optimal Transport Conditional Flow Matching (OT-CFM) \cite{lipman2023flow}. Given a base sample $x_0 \sim \mathcal{N}(0,I)$ and a data sample $x_1 \sim p_{\mathrm{data}}$, OT-CFM trains a time-dependent vector field $v_\theta(x,t)$ by minimizing
\[
\mathcal{L}_{\mathrm{FM}}(\theta)
=
\mathbb{E}_{t,x_0,x_1}
\left[
\left\|
v_\theta(x_t,t) - u_t
\right\|_2^2
\right],
\]
where $t\sim \mathcal{U}(0,1)$, $x_t=(1-t)x_0+t x_1$, and $u_t=x_1-x_0$ for the deterministic linear interpolation path. In our implementation, pairs $(x_0,x_1)$ were coupled using minibatch optimal transport. Our process model is a 34.5 million parameter UNet architecture. OT-CFM regresses a vector field between a simple Gaussian base distribution and the empirical data distribution using deterministic optimal transport plans. This approach provides highly stable training dynamics and bypasses the slow iterative sampling bottlenecks characteristic of standard diffusion models. During inference we used Dormand-Prince sampling \cite{Dormand1980AFO} via the \texttt{torchdiffeq} package \cite{torchdiffeq}. We generated a candidate pool of 20,000 synthetic slices, which were subsequently routed to the supervisor for filtering to get our final set of privacy focused synthetic images.

\subsection{Geometric Latent Embedding Spaces}
Pixel-level distances are poorly aligned with the privacy risks in lung CT slices: two images can differ in intensity or small local texture while preserving the same patient-specific anatomical structure. To obtain more meaningful privacy scores, the Supervisor embeds both generated candidates and training-reference slices into learned latent spaces produced by auxiliary autoencoders trained on the same image domain. These encoders are not used to generate images; they are used only to score and filter candidate samples after generation. Note that for each encoder, the decoder is never used during inference and only used during encoder training.

\begin{itemize}
    \item \textbf{Spatial Autoencoder:}
    The spatial autoencoder is a convolutional encoder-decoder trained to reconstruct 256$\times$256 grayscale CT slices. Its encoder compresses each image into an 8-channel spatial latent tensor at 32$\times$32 resolution, preserving coarse anatomical layout while discarding pixel-level noise. For the spatial Supervisor, these latent tensors are flattened and L2-normalized, and each generated image is scored by cosine similarity to its nearest training slice. Higher distance from the nearest training latent is treated as a stronger privacy signal.

    \item \textbf{Contrastive Autoencoder:}
    The contrastive autoencoder uses the same reconstruction backbone as the spatial autoencoder, but adds an auxiliary NT-Xent contrastive objective \cite{agren2022ntxentlossupperbound} over CT-safe augmented views. This encourages the encoder to organize slices by stable anatomical structure rather than reconstruction details alone. The contrastive approach was motivated by the need for a latent space that is regularized enough to use a manifold geodesic. In the geodesic supervision, we use the trained encoder's spatial latents, pool them into compact descriptors, apply PCA whitening, and build a $k$-nearest-neighbor graph over the training-reference manifold. Generated images are then scored by graph-based geodesic distance to the training bank, which penalizes candidates that lie close to dense regions associated with real training anatomy.

    \item \textbf{Riemannian Autoencoder:}
    The Riemannian encoder is a spatial autoencoder with an added metric head layer. The encoder first maps each CT slice \(x_i\) to a spatial latent tensor \(z_i\). The metric head then pools this latent tensor, flattens it, and passes it through a small MLP to produce a compact 128-dimensional embedding:
    
    \[
    z_i = E_\theta(x_i), \qquad h_i = M_\phi(z_i).
    \]
    
    During training, augmented views of the same slice are treated as positives. For each embedding, the model estimates local PCA geometry from nearby embeddings, producing tangent directions and local variances. Distances are then measured anisotropically:
    
    \[
    d_{\mathrm{RAM}}^2(i,j)
    =
    \alpha d_{\parallel}^2(i,j)
    +
    \beta d_{\perp}^2(i,j),
    \]
    
    where \(d_{\parallel}\) measures displacement along local tangent directions and \(d_{\perp}\) measures displacement away from the local manifold.
    
    This distance is used in a contrastive loss so the metric head learns an embedding space where distances reflect local anatomical geometry. At inference, the Supervisor uses the frozen encoder and metric head to embed generated and training images, then deprioritizes generated samples that are too close to the training manifold under this Riemannian distance.
    
\end{itemize}

\subsection{Post-Generation Selection and Privacy Gating}

Rather than constraining the generator during sampling, our pipeline separates generation from privacy filtering. The flow-matching model first generates an unconstrained candidate pool $\mathcal{G}=\{g_i\}_{i=1}^{N}$, with $N=20{,}000$ for our main runs. The Supervisor then scores each generated slice against the training-reference bank $\mathcal{T}$ in one of several learned latent spaces. Candidates that are farther from the training bank are assigned higher privacy scores and are preferred during subset selection.

\paragraph{Spatial cosine gate.}
For the spatial gate, each image is encoded with the spatial autoencoder encoder $E_s$. The spatial latent is flattened and L2-normalized:
\[
    \phi_s(x)
    =
    \frac{\mathrm{vec}(E_s(x))}
    {\|\mathrm{vec}(E_s(x))\|_2}.
\]
For a generated image $g$, we compute its nearest training similarity
\[
    s_{\max}(g)
    =
    \max_{t \in \mathcal{T}}
    \phi_s(g)^\top \phi_s(t).
\]
The spatial privacy score is the corresponding cosine distance:
\[
    q_{\mathrm{spatial}}(g)
    =
    \max(0, 1 - s_{\max}(g)).
\]
Thus, generated images whose latent representation is very close to a training slice receive lower privacy scores.

\paragraph{Geodesic gate.}
For the geodesic gate, we use the contrastively trained encoder to obtain spatial latents, pool them to a fixed grid, and flatten them into descriptors $p(x)$. A PCA-whitening transform is fit on the training descriptors:
\[
    y(x)
    =
    \left(
        \frac{p(x)-\mu}{\sigma}
    \right)
    V_r^\top \Lambda_r^{-1/2},
\]
where $\mu$ and $\sigma$ are the training descriptor mean and standard deviation, and $V_r,\Lambda_r$ are the retained PCA directions and variances.

We then build a $k$-nearest-neighbor graph over the union of training and generated descriptors. A local density estimate is computed from the $k_d$ nearest training descriptors:
\[
    \rho_i
    =
    \frac{1}{k_d}
    \sum_{m=1}^{k_d}
    \exp\left(
        -\frac{\|y_i-y_{n_m(i)}\|_2^2}{2\sigma_\rho^2}
    \right).
\]
Edges are weighted by density-scaled squared distance:
\[
    w_{ij}
    =
    \|y_i-y_j\|_2^2
    \cdot
    \frac{1}{2}
    \left(
        \rho_i^{-1}+\rho_j^{-1}
    \right).
\]
A multi-source shortest-path search is initialized from all training nodes. The geodesic privacy distance for a generated candidate $g$ is
\[
    d_{\mathrm{geo}}(g)
    =
    \sqrt{
        \min_{\pi:g \rightarrow \mathcal{T}}
        \sum_{(i,j)\in\pi} w_{ij}
    }.
\]
Larger geodesic distance indicates greater separation from the training manifold and therefore a higher privacy score.

\paragraph{Riemannian Anti-Memory (RAM) privacy gate.}

Intuitively, RAM is an implicit manifold construction inspired by moving least squares style approximation in higher dimensions. The Riemannian encoder maps each image to a compact metric embedding $h(x)$. For each training embedding $h_j$, we fit a local PCA geometry over its $k_n$ nearest training neighbors. This gives a local mean, tangent basis $U_j$, tangent eigenvalues $\lambda_j$, and normal variance $\sigma_{\perp,j}^2$.

For a generated embedding $h(g)$ and a nearby training embedding $h_j$, define
\[
    \Delta_j = h(g)-h_j.
\]
The local anisotropic distance is
\[
    d_{\mathrm{RAM}}^2(g,j)
    =
    \sum_{\ell=1}^{r}
    \frac{
        (u_{j,\ell}^{\top}\Delta_j)^2
    }{
        \lambda_{j,\ell}+\eta
    }
    +
    \frac{
        \|\Delta_j-U_jU_j^\top\Delta_j\|_2^2
    }{
        \sigma^2_{\perp,j}+\eta
    }.
\]
The RAM gate assigns each candidate its nearest valid training-patch distance:
\[
    d_{\mathrm{RAM}}^2(g)
    =
    \min_{j \in \mathcal{N}_{k_q}(g)}
    d_{\mathrm{RAM}}^2(g,j),
\]
where $\mathcal{N}_{k_q}(g)$ is a shortlist of nearest training embeddings, $r$ is the number of retained tangent directions, and $k_q$ is the number of nearest training embeddings considered when computing the candidate's minimum RAM distance. The constants $\alpha$ and $\beta$ weight tangent and normal displacement during metric-head training. $\eta$ is a small numerical stability constant. During inference, all local PCA bases and variances are computed only from the training embeddings and then held fixed. Candidates with small RAM distance are considered higher-risk because they lie close to local training anatomy under the learned anisotropic metric. 

\paragraph{Selection Methods.}
Following the privacy scoring phase, the candidate pool must be reduced to the challenge requirement of 5,000 images. To prevent mode collapse and ensure the final synthetic dataset reflects a wide distribution of patient anatomies, we bypass naive greedy sorting. Instead, we employ coreset selection strategies. Specifically, we utilize Determinantal Point Processes (DPP) \cite{kulesza2012determinantal} to probabilistically select subsets that maximize the spanned feature volume. We also apply Stein Kernel Thinning (SKT) \cite{dwivedi2021kernel}, leveraging a score function to systematically compress the candidate distribution while maintaining data diversity. For DPP selection, we used the Supervisor embedding kernel with privacy scores as quality weights; for SKT, thinning was performed in the same embedding space using the corresponding kernel score over the 20,000-candidate pool.

\subsection{Internal Privacy and Utility Metrics During Development}

We built an internal evaluation suite to approximate the major failure modes we expected: poor visual fidelity, distributional mismatch, direct memorization of training slices, and excessive similarity between generated images and the training set. These metrics were only used during offline model evaluation.

Let $\mathcal{T}$ denote the training split, $\mathcal{H}$ an internal holdout split, and $\mathcal{G}$ the generated sample set. We chose an 80/20 train/holdout split and the same split was used for all training runs. We evaluated each submission candidate in medical feature spaces extracted with BioMedCLIP \cite{zhang2024biomedclip} and RadImageNet/InceptionV3 \cite{radimagenet}.

For distributional metrics, we used a triangulated protocol with three comparisons:
\[
    \text{baseline}: \mathcal{T} \leftrightarrow \mathcal{H},
    \qquad
    \text{generalization}: \mathcal{G} \leftrightarrow \mathcal{H},
    \qquad
    \text{memorization}: \mathcal{G} \leftrightarrow \mathcal{T}.
\]
The baseline comparison estimates the natural train--holdout gap. The generalization comparison measures whether generated samples resemble unseen real CT slices, while the memorization comparison helps identify whether generated samples are unusually close to the training distribution.

\paragraph{Feature-space FID.}
We computed the FID in both BioMedCLIP and RadImageNet feature spaces. For two embedding sets $A$ and $B$ with empirical means $\mu_A,\mu_B$ and covariances $\Sigma_A,\Sigma_B$, we used
\[
    \mathrm{FID}(A,B)
    =
    \|\mu_A-\mu_B\|_2^2
    +
    \mathrm{Tr}
    \left(
        \Sigma_A+\Sigma_B
        -
        2(\Sigma_A\Sigma_B)^{1/2}
    \right).
\]
Low generated--holdout FID indicates visual and anatomical realism. However, a generated--training FID that is much better than the train--holdout baseline can indicate overfitting or memorization.

\paragraph{PRDC density and coverage.}

To complement FID, we computed PRDC-style density and coverage metrics over normalized medical embeddings \cite{prdc}. Coverage measures the fraction of real holdout samples whose local neighborhood contains at least one generated sample, while density measures how many generated samples fall within real-data neighborhoods. These metrics helped distinguish realistic but low-diversity outputs from outputs that covered a broader range of lung CT anatomy.

\paragraph{Maximum mean discrepancy.}
We also used an  unbiased RBF-MMD$^2$ \cite{mmd} over normalized embeddings:
\[
    \mathrm{MMD}^2(A,B)
    =
    \underbrace{\frac{1}{m(m-1)}\sum_{i\ne j}k(a_i,a_j)}_{\text{within}\  A\ \text{similarity}}
    +
    \underbrace{\frac{1}{n(n-1)}\sum_{i\ne j}k(b_i,b_j)}_{\text{within}\  B\ \text{similarity}}
    -
    2 \cdot \underbrace{\frac{1}{mn}\sum_{i,j}k(a_i,b_j)}_{\text{between-set similarity}},
\]
where $k$ is an RBF kernel and the bandwidth was calibrated from holdout feature distances. MMD was useful as a second distributional test because it is sensitive to differences in the full embedding distribution, not only the first two moments used by FID.

\paragraph{Nearest-neighbor (NN) privacy checks.}
To approximate instance-level memorization, we computed NN privacy metrics between $\mathcal{G}$ and $\mathcal{T}$. For each generated image, we first found a shortlist of nearest training candidates in medical embedding space. We then refined these candidate pairs using image-space similarity metrics: MS-SSIM \cite{msssim}, and LPIPS \cite{lpips}. A generated image was considered more concerning when it had unusually high MS-SSIM or unusually low LPIPS to a training image.

Because real CT slices from the same distribution can naturally be similar, we calibrated near-duplicate thresholds against train--train self-neighbor statistics. For example, generated samples were flagged when their best SSIM exceeded a high percentile of the train self-neighbor SSIM distribution, or when their best LPIPS fell below a low percentile of the train self-neighbor LPIPS distribution.

\section{Official Results}
The performance of our generator configurations and post-hoc Supervisor filtering mechanisms are presented in Table \ref{tab:results}. All submissions were evaluated by the ImageCLEFmed organizers using a held-out patient dataset to act as a negative control. The primary metrics are Fr\'{e}chet Inception Distance (FID) to measure visual realism, and the composite Privacy Preservation Score (PPS), which is derived from the four independent privacy attack leakage scores ($L_1$--$L_4$).

\subsection{Evaluation Protocol}
It is important to note that the specific methodologies of the privacy attacks were not disclosed to participants prior to the submission deadline. Our approached were engineered from first principles of geometric filtering rather than optimized against known scoring functions. Post-submission, the organizers evaluated the synthetic datasets against a held-out negative control set using four distinct attack vectors \cite{ImageCLEFmedicalGANs2026}:

\begin{itemize}
    \item \textbf{Attack 1 (Nearest-Neighbour Distance Audit):} Measures instance-level memorization by calculating the feature-space distance between synthetic images and their nearest training counterparts.
    \item \textbf{Attack 2 (Membership Inference Attack):} Evaluates distributional bias by using an AUC-ROC classifier to determine if the synthetic distribution reveals which specific patients were used in the training set.
    \item \textbf{Attack 3 (Patient Re-identification):} Assesses deep structural identity leakage by building per-patient anatomical centroids and measuring the rank-1 patient coverage of the synthetic images.
    \item \textbf{Attack 4 (Generalized Inversion Attack):} A white-box attack measuring latent-space memorization by utilizing gradient-based optimization to reconstruct target real images from the generator's checkpoint.
\end{itemize}

These four leakage scores ($L_1$--$L_4$) are averaged and inverted to calculate the final composite Privacy Preservation Score (PPS) as

\[
\mathrm{PPS} = 1 - \frac{1}{4} \sum_{i=1}^4 L_i.
\]

\begin{table*}[t]
\centering
\caption{Official ImageCLEFmed GANs 2026 Evaluation Results. Methods are evaluated on visual fidelity (FID) and Privacy Preservation Score (PPS). The composite PPS is derived from four distinct privacy attacks ($L_1$--$L_4$). \textbf{Bold} values indicate the best performance in each column and the \colorbox{gray!15}{highlighted} line indicates the best model.}
\label{tab:results}
\begin{tabular}{lcccccc}
\toprule
\textbf{Method} & \textbf{FID} ($\downarrow$) & \textbf{PPS} ($\uparrow$) & \textbf{Attack 1} ($\downarrow$) & \textbf{Attack 2} ($\downarrow$) & \textbf{Attack 3} ($\downarrow$) & \textbf{Attack 4} ($\downarrow$) \\
\midrule
\texttt{fm-unet-300}          & 0.3385 & 0.358 & 0.1459 & 0.5916 & 0.9933 & 0.837 \\
\texttt{sv-spatial-dpp-300}        & 0.5495 & 0.380 & 0.1124 & 0.4902 & 0.9933 & 0.887 \\
\texttt{sv-spatial-kt-300}         & 0.5645 & 0.449 & 0.1103 & 0.4775 & 0.9933 & 0.627 \\
\texttt{sv-geodesic-wdpp-300}      & 0.6355 & 0.450 & 0.1161 & 0.5180 & 1.0000 & 0.565 \\
\texttt{sv-geodesic-skt-300}       & 0.9542 & 0.380 & 0.1040 & 0.4678 & 1.0000 & 0.907 \\
\midrule
\rowcolor{gray!25}
\texttt{fm-unet-100}            & 0.3290 & \textbf{0.549} & 0.0901 & 0.3797 & 0.9933 & \textbf{0.341} \\
\texttt{sv-spatial-dpp-100}     & \textbf{0.2639} & 0.492 & 0.0835 & \textbf{0.3519} & \textbf{0.9732} & 0.625 \\
\texttt{sv-geodesic-greedy-100} & 0.4272 & 0.498 & \textbf{0.0804} & 0.3725 & 0.9866 & 0.562 \\
\texttt{sv-riemannian-wdpp-100}    & 0.3250 & 0.466 & 0.0880 & 0.3838 & 0.9933 & 0.671 \\
\texttt{sv-riemannian-skt-100}     & 0.3758 & 0.433 & 0.0874 & 0.3724 & 0.9933 & 0.814 \\
\bottomrule
\end{tabular}
\end{table*}

\section{Discussion}

The results reveal several counter-intuitive trade-offs between generative realism and patient privacy. By analyzing the performance of our multi-layered defense mechanisms across the four distinct attack vectors, several insights emerge that possibly challenge standard assumptions in privacy-preserving machine learning.

\subsection{Training Duration and the Privacy--Utility Trade-off}

The official results suggest that training duration played an important role in the balance between visual fidelity and empirical privacy. The model trained for 300 epochs (\texttt{fm-unet-base}) achieved an FID of 0.3385 and a Privacy Preservation Score (PPS) of 0.358, whereas the shorter 100-epoch run (\texttt{fm-unet-100}) achieved a slightly better FID of 0.3290 and a substantially higher PPS of 0.549. The 100-epoch model also obtained lower leakage scores for nearest-neighbor memorization, membership inference, and generalized inversion, while patient re-identification remained effectively unchanged. Applying spatial Determinantal Point Process filtering to samples from the 100-epoch model further improved visual fidelity, producing the best FID among our submissions at 0.2639 while retaining a PPS of 0.492. These findings suggest that extending training to 300 epochs did not improve realism and was associated with increased empirical privacy leakage, potentially because prolonged optimization allowed the model to capture increasingly specific characteristics of the training data; however, because the two runs were not part of a controlled training-duration ablation, this relationship should be interpreted as an observed association rather than a definitive causal effect.

\subsection{Efficacy of Geometric Latent Filtering}
Our inference-time Supervisor pipeline was robust against instance-level memorization (Attack 1) and distributional bias (Attack 2). By over-generating candidate images and passing them through geometric latent gates, we successfully penalized 1-to-1 visual regurgitation of the training data. Runs utilizing our subset selection strategies \cite{kulesza2012determinantal, dwivedi2021kernel}, such as \texttt{sv-geodesic-greedy-100}, drove the Attack 1 leakage score down to a remarkably low 0.0804. This demonstrates that mapping generated samples to custom-trained spatial and contrastive autoencoder spaces, followed by rigorous coreset selection, is a robust solution for reducing explicit visual memorization.

\subsection{Resilience to White-Box Inversion (Attack 4)}
A notable result in our evaluation is the strong performance of the
unfiltered \texttt{fm-unet-100} model against the generalized inversion
attack (Attack 4). While the Supervisor pipeline was most effective
against instance-level visual memorization in Attacks 1 and 2,
\texttt{fm-unet-100} achieved an Attack 4 leakage score of 0.341,
substantially outperforming its post-hoc filtered variants.

This result is likely explained by the shorter training schedule.
\texttt{fm-unet-100} was trained for 100 epochs, compared with 300 epochs
for \texttt{fm-unet-300}. Limiting training to 100 epochs reduced the
amount of training-specific information encoded in the generator's
parameters, making reconstruction through white-box inversion more
difficult. This finding demonstrates that shorter training duration can reduce
parameter-level memorization, while geometric filtering primarily reduces
direct visual similarity to training images. The filtered variants used
the same 100-epoch generator together with an additional Supervisor
autoencoder, whose inclusion in the submitted artifacts may explain their
higher Attack 4 leakage scores.

\subsection{The Reality of Identity Leakage (Attack 3)}

While our framework provides some protection against verbatim visual copying, the evaluation exposed a critical vulnerability regarding deep structural identity. Across all submissions—including our most aggressively filtered models—the leakage score for Patient Re-identification (Attack 3) remained near maximum, with rank-1 patient coverages hovering between 0.97 and 1.00. 

Initially, this metric appears counter-intuitive; broad representation across the generated dataset might mistakenly be interpreted as healthy diversity rather than mode collapse. However, in the context of privacy, a 1.0 coverage metric indicates a near-total leakage under this rank-1 re-identification metric. This suggests the generated distribution may retain patient-specific anatomical structure beyond what would be expected from generic anatomy alone.

This finding presents a vital realization for the field of secure medical generation: while inference-time geometric filtering and subset selection help enforce visual diversity, they cannot scrub the fundamental patient identities encoded in the generator’s learned representation during optimization. Future architectures must address this disparity, acknowledging that defeating nearest-neighbor visual attacks does not equate to protecting underlying anatomical identity.

\subsection{Internal Metrics Results}

The results of our internal evaluation metrics can be found in the Appendix.

\section{Future Work}

A significant challenge encountered during the development of our evaluation pipeline was the abstract nature of patient ``fingerprints.'' Prior to the release of the final evaluation metrics, our team lacked a standardized, mathematical definition of what separates a unique anatomical fingerprint of lung CT scans from generalized human anatomy. Consequently, engineering local loss functions or distance metrics to penalize this deep structural leakage without a formal target proved to be difficult. 

Future research must focus on formally quantifying these identity-defining features (such as specific bone density ratios, highly individualized organ topologies, or vascular network geometries) into standardized, differentiable metrics. One approach pursued but not completed in time was to apply Source Guided Flow Matching (SGFM) \cite{sgfm} to be able to select samples from the input noise distribution that would facilitate generating (and avoiding) specific anatomies. Our model was trained in the full pixel-space. Training in a lower-dimensional latent space could allow for more stable convergence and potentially permit the use of techniques such as differential privacy.

\section{Conclusion}

In this work, we presented a two-stage, privacy-focused framework for synthetic medical image generation developed for the ImageCLEFmed GANs 2026 challenge. Our architecture combines an Optimal Transport Conditional Flow Matching generator with an inference-time Supervisor pipeline for geometric filtering and subset selection.

Our evaluation showed that the shorter 100-epoch run (\texttt{fm-unet-100}) provided a more favorable balance between visual fidelity and empirical privacy than the 300-epoch baseline, achieving our highest Privacy Preservation Score of 0.549 with an FID of 0.3290. Applying spatial DDP selection to this run produced our best FID of 0.2639 while retaining a comparatively strong PPS of 0.492. Geometric filtering generally reduced nearest-neighbor memorization and membership-inference leakage, as measured by Attacks 1 and 2, although its effect on the generalized inversion attack was mixed.

However, the persistence of high patient re-identification leakage under Attack 3 across all submissions highlights an important challenge for privacy-focused generative modeling. Future approaches must move beyond preventing direct image copying and address the retention of patient-specific anatomical structure while preserving the broader clinical characteristics required for realistic and useful synthetic datasets.

\section*{Acknowledgments}

We thank the Data Science at Georgia Tech (DS@GT) CLEF competition group for their support. This research was supported in part through research cyberinfrastructure resources and services provided by the Partnership for an Advanced Computing Environment (PACE) at the Georgia Institute of Technology, Atlanta, Georgia, USA \cite{PACE}. We  also thank the organizers of the CLEF conference and competition.

\section*{Declaration on Generative AI}

During the preparation of this work, the authors used Google Gemini and ChatGPT to help write LaTeX code. OpenAI's Codex and Anthropic's Claude models were also used during the software development phase of the work. After using these tools, the authors reviewed and edited the content as needed and take full responsibility for the publication's content. 

\section{Appendix}

Below we report our internal evaluation results, which were used during development before the official scoring criteria were released. Since the organizers indicated that privacy, realism, and diversity would be evaluated, we selected model, supervisor, and selection variants that spanned this trade-off. The supervisor encoders were not trained with DP, as stable DP convergence was achieved only near the submission deadline.

A notable result is that the \texttt{sv-riemannian-100} models improved privacy with little change in RadImageNet metrics. Since RadImageNet captures high-frequency radiological texture while BioMedCLIP captures higher-level semantic structure, this suggests improved privacy with limited loss of radiological utility. LPIPS$_{\min}$ should therefore be interpreted together with the feature space: low values may indicate local similarity to training images, while RadImageNet stability suggests preserved texture-scale detail. The ideal scenario would be a low LPIPS$_{\min}$ for RadImageNet and a high LPIPS$_{\min}$ for BioMedCLIP.

\begin{table}[h]
\centering
\scriptsize
\setlength{\tabcolsep}{3.5pt}
\caption{Generated embedding cosine distance delta, memorization MMD$^2$, and LPIPS$_{\min}$ across feature spaces. \underline{Underlined} values indicate the best value within each method block, while \textbf{bold} values indicate the best value across the full column.}
\label{tab:combined_cosine_mmd_lpips}
\begin{tabular}{lcccccc}
\toprule
& \multicolumn{3}{c}{\textbf{RadImageNet Feature Space}} 
& \multicolumn{3}{c}{\textbf{BioMedCLIP Feature Space}} \\
\cmidrule(lr){2-4} \cmidrule(lr){5-7}
\textbf{Method} 
& \textbf{Cosine $\Delta$ (\%) $\downarrow$} 
& \textbf{Mem. MMD$^2$ $\downarrow$} 
& \textbf{LPIPS$_{\min}$ $\downarrow$} 
& \textbf{Cosine $\Delta$ (\%) $\uparrow$} 
& \textbf{Mem. MMD$^2$ $\downarrow$} 
& \textbf{LPIPS$_{\min}$ $\uparrow$} \\
\midrule
\texttt{fm-unet-300} & 0.00 & 0.00737 & \textbf{0.12602} & 0.00 & 0.05256 & 0.13063 \\
\midrule
\texttt{sv-geodesic-dpp-100} & +18.24 & 0.00473 & 0.19137 & +25.70 & \underline{0.02950} & 0.19706 \\
\texttt{sv-geodesic-greedy-100} & +33.04 & 0.00703 & 0.19887 & \textbf{+34.83} & 0.04335 & \textbf{\underline{0.20530}} \\
\texttt{sv-geodesic-kt-100} & \underline{+12.92} & \underline{0.00410} & \underline{0.18754} & +20.99 & 0.03184 & 0.19380 \\
\texttt{sv-geodesic-skt-100} & +33.04 & 0.00703 & 0.19887 & \textbf{+34.83} & 0.04335 & 0.20528 \\
\texttt{sv-geodesic-wdpp-100} & +18.72 & 0.00473 & 0.19160 & +26.05 & 0.03053 & 0.19719 \\
\midrule
\texttt{sv-riemannian-dpp-100} & +1.31 & 0.00402 & 0.17978 & \underline{+15.60} & 0.02939 & 0.18466 \\
\texttt{sv-riemannian-greedy-100} & +0.52 & 0.00511 & 0.18023 & +13.36 & 0.02902 & 0.18438 \\
\texttt{sv-riemannian-kt-100} & \textbf{-1.76} & 0.00498 & \underline{0.17744} & +10.64 & \underline{0.02564} & 0.18279 \\
\texttt{sv-riemannian-skt-100} & +0.47 & 0.00502 & 0.18020 & +13.36 & 0.02928 & 0.18434 \\
\texttt{sv-riemannian-wdpp-100} & +1.75 & \textbf{0.00354} & 0.18028 & +15.59 & 0.02688 & \underline{0.18488} \\
\midrule
\texttt{sv-spatial-dpp-100} & +17.22 & 0.00473 & 0.19655 & \underline{+25.24} & \textbf{0.02517} & 0.20316 \\
\texttt{sv-spatial-greedy-100} & +19.75 & 0.00618 & 0.19699 & +21.27 & 0.03681 & 0.20153 \\
\texttt{sv-spatial-kt-100} & \underline{+7.10} & \underline{0.00461} & \underline{0.18707} & +15.06 & 0.02984 & 0.19244 \\
\texttt{sv-spatial-skt-100} & +19.79 & 0.00599 & 0.19701 & +21.30 & 0.03407 & 0.20154 \\
\texttt{sv-spatial-wdpp-100} & +18.66 & 0.00482 & 0.19732 & +25.12 & 0.02882 & \underline{0.20361} \\
\midrule
\texttt{sv-geodesic-dpp-300} & +31.25 & \underline{0.00897} & \underline{0.15079} & +18.63 & 0.07414 & 0.15553 \\
\texttt{sv-geodesic-greedy-300} & +49.56 & 0.01585 & 0.16325 & \underline{+26.85} & 0.10010 & 0.16887 \\
\texttt{sv-geodesic-kt-300} & \underline{+24.10} & 0.01020 & 0.15366 & +15.95 & \underline{0.06932} & 0.15762 \\
\texttt{sv-geodesic-skt-300} & +49.63 & 0.01594 & 0.16325 & \underline{+26.85} & 0.09872 & \underline{0.16889} \\
\texttt{sv-geodesic-wdpp-300} & +32.66 & 0.00925 & 0.15129 & +19.46 & 0.07661 & 0.15615 \\
\midrule
\texttt{sv-riemannian-dpp-300} & +8.49 & 0.00761 & 0.13526 & +5.93 & 0.06512 & 0.13979 \\
\texttt{sv-riemannian-greedy-300} & +10.57 & 0.00794 & 0.13817 & +4.00 & 0.08219 & 0.14163 \\
\texttt{sv-riemannian-kt-300} & \underline{+8.10} & 0.00723 & \underline{0.13488} & +2.98 & \underline{0.06317} & 0.13902 \\
\texttt{sv-riemannian-skt-300} & +10.57 & 0.00819 & 0.13819 & +4.05 & 0.07222 & \underline{0.14164} \\
\texttt{sv-riemannian-wdpp-300} & +9.40 & \underline{0.00710} & 0.13662 & \underline{+5.96} & 0.06503 & 0.14062 \\
\midrule
\texttt{sv-spatial-dpp-300} & +35.38 & 0.00993 & 0.15704 & +21.16 & 0.07519 & 0.16416 \\
\texttt{sv-spatial-greedy-300} & +34.54 & 0.01092 & 0.16497 & +14.55 & 0.08294 & 0.16849 \\
\texttt{sv-spatial-kt-300} & \underline{+18.62} & \underline{0.00847} & \underline{0.15181} & +8.78 & \underline{0.07489} & 0.15534 \\
\texttt{sv-spatial-skt-300} & +34.55 & 0.01076 & 0.16493 & +14.71 & 0.08701 & \underline{0.16854} \\
\texttt{sv-spatial-wdpp-300} & +37.22 & 0.00979 & 0.15962 & \underline{+21.37} & 0.07767 & 0.16633 \\
\bottomrule
\end{tabular}
\end{table}

\pagebreak

\bibliography{main}

\end{document}